\documentclass[sigconf]{acmart}
\usepackage{graphicx}    
\usepackage{subcaption}  
\usepackage{hyperref}    

\acmConference[SIGSPATIAL '25]{ACM International Conference on Advances in Geographic Information Systems, November 3-November 6, 2025, Minneapolis, MN, USA}

\begin{document}

\title{Improving the Computational Efficiency and Explainability of GeoAggregator}

\author{Rui Deng}
\affiliation{%
  \institution{University of Glasgow}
  \city{Glasgow}
  \country{United Kingdom}}
\email{rui.deng@glasgow.ac.uk}

\author{Ziqi Li}
\affiliation{%
  \institution{Florida State University}
  \city{Tallahassee}
  \country{United States}}
\email{ziqi.li@fsu.edu}

\author{Mingshu Wang}
\affiliation{%
  \institution{University of Glasgow}
  \city{Glasgow}
  \country{United Kingdom}}
\email{mingshu.wang@glasgow.ac.uk}

\renewcommand{\shortauthors}{Rui et al.}


\begin{abstract}
Accurate modeling and explaining geospatial tabular data (GTD) are critical for understanding geospatial phenomena and their underlying processes. Recent work has proposed a novel transformer-based deep learning model named GeoAggregator (GA) for this purpose, and has demonstrated that it outperforms other statistical and machine learning approaches. In this short paper, we further improve GA by 1) developing an optimized pipeline that accelerates the data-loading process and streamlines the forward pass of GA to achieve better computational efficiency; and 2) incorporating a model ensembling strategy and a post-hoc model explanation function based on the GeoShapley framework to enhance model explainability. We validate the functionality and efficiency of the proposed strategies by applying the improved GA model to synthetic datasets. Experimental results show that our implementation improves the prediction accuracy and inference speed of GA compared to the original implementation. Moreover, explanation experiments indicate that GA can effectively captures the inherent spatial effects in the designed synthetic dataset. The complete pipeline has been made publicly available for community use (\url{https://github.com/ruid7181/GA-sklearn}).
\end{abstract}




\maketitle

\section{INTRODUCTION}
Geospatial tabular data (GTD) is increasingly available with the widespread deployment of GPS-enabled sensors and the growing demand for location-based services \cite{wu2024geospatial}. Composed of geo-referenced sample points (rows) and mixed types of features (columns), GTD represents a specialized data modality beyond traditional tabular data. The understanding of GTD involves various geospatial tasks, including predictive modeling \cite{dai2022geographically}, process explanation \cite{li2024geoshapley}, and even uncertainty assessment \cite{wu2024geospatial}. These tasks are crucial in natural and social sciences to understand intricate spatial relationships, predict spatial scenarios, and inform decision-making.

Inspired by the success of deep learning in other AI tasks, many efforts have been made in recent years to extend deep learning approaches to GTD-related tasks, with examples covering a wide range of network architectures, such as dense networks \cite{Du-GNNWR}, graph-based models \cite{zhu2022spatial}, convolutional networks \cite{dai2022geographically} and transformers \cite{deng2025geoaggregator}. Although these models exhibit promising performance advantages, they are often associated with a large number of parameters and require considerable computational resources. This has limited their scalability and usability in handling large GTDs in empirical applications. Nevertheless, the black box nature of most machine learning models poses challenges in drawing meaningful insights into how different geospatial phenomena interact with each other, and whether the models are trustworthy in real-world applications.

One advancement in deep GTD modeling is the recently proposed GeoAggregator (GA) model \cite{deng2025geoaggregator}. Designed on the basis of the transformer architecture, GA exhibits strong performance with highlighted computational efficiency. The model learns directly from the GTD without the need to build additional proxy grids or graph structures. This has enhanced the flexibility and efficiency of the model, enabling it to capture more complex spatial interactions. These designs allow GA to scale to large GTDs with linearly growing computational cost. Despite its theoretical linear complexity, the original implementation of GA (\verb|GA-sklearn|) has a bottleneck related to data loading operations. Addressing this will further enhance its empirical efficiency.

In addition to the model efficiency, explaining deep models has gained great attention in geospatial researches \cite{roussel2023geospatial, liu2024explainable, li2024geoai}. One of the most popular explanation approaches is SHapley Additive exPlanation (SHAP) \cite{NIPS2017_7062}. Originated from game theory \cite{shapley1953value}, this method calculates the Shapley value for each feature, which reflects the marginal contribution to the final prediction. Internally, Shapley values are calculated based on predictions from different subsets of the features. To date, several approaches have been proposed to efficiently estimate SHAP values of different mainstream machine learning models, including Tree SHAP, Kernel SHAP, and Deep SHAP \cite{NIPS2017_7062}. Recently, GeoShapley, an extension of the Shapley value approach for geo-referenced data was developed to measure geospatial effects and the interaction between spatial and non-spatial features \cite{li2024geoshapley}. Nevertheless, few works have provided SHAP-based explanations of sophisticated geospatial deep learning models. Here, we leverage the GeoShapley explanation framework to address this gap.

To summarize, in this paper, we integrate the GA model in an optimized computational pipeline that further improves the accuracy and computational efficiency of the model. We demonstrate the improved performance of GA by comparing the new implementation with the original and a strong baseline model on a series of synthetic datasets. Using the proposed pipeline, we also explore the explainability of GA using the GeoShapley framework. Our implementation has been made publicly available for users to perform geospatial prediction and analysis that involve GTD understanding tasks \cite{ruid7181-ga-sklearn}.

\section{METHODOLOGY AND IMPLEMENTATION}
In this section, we first review the technical designs of the GA model based on the original implementation and propose optimization strategies in an attempt to further boost the prediction accuracy and model explainability.

\subsection{GeoAggregator Principles}
GA is a transformer-based deep learning model designed specifically for supervised geospatial regression tasks. Unlike many deep learning alternatives for the spatial regression task, GA does not rely on predefined spatial graph structures or proxy grids to capture contextual knowledge. Instead, it learns directly from related rows of GTD, taking an arbitrarily long contextual point sequence of a target point as input, at each forward pass. To better capture various spatial effects such as Spatial Autocorrelation (SA) and Spatial Heterogeneity (SH), GA leverages local attention and rotary positional embedding to flexibly adapt to the heterogeneous spatial effects in both covariates and the target variable. Furthermore, taking inspiration from the Geographically Weighted Regression (GWR) model, GA explicitly adjusts the attention weights in the transformer using a Gaussian spatial kernel. For a target point ${\boldsymbol{p}_i}$ and a set of its spatial contextual points ${\mathcal{N}_i = \{\boldsymbol{p}_j\}}$ (including the target point itself), the updated feature ${\boldsymbol{p}_i}$ through the multi-head attention is given by:

\begin{equation}
    \alpha_{ij} = \text{softmax}((\mathbf{W}^Q \mathbf{e}_i)^\top \mathbf{W}^K \mathbf{e}_j - \lambda d^{2}_{ij})
\end{equation}

\begin{equation}
    \mathbf{e}_i = \sum_{j} \alpha_{ij} \mathbf{W}^V \mathbf{e}_j
\end{equation}

\begin{equation}
    \tilde{\mathbf{e}}_i = \mathbf{W}^O \cdot \mathbf{e}_i
\end{equation}
where ${\mathbf{e}_i}$ is a feature representation of ${\boldsymbol{p}_i}$, ${\mathbf{W}^{\{Q, K, V, O\}}}$ are learnable projection matrices, and ${\lambda}$ is a learnable attention bias factor (ABF) that controls the magnitude of the added Gaussian bias.

In the original implementation (equations (1) to (3)), the multi-head attention mechanism only allows one ABF across all attention heads. This could lead to suboptimal solutions because a single ABF could limit the multi-head attention mechanism from learning multiple patterns at different scales. To this end, here we assign one learnable ABF (${\lambda^{(h)}}$) for the ${h}$-th attention head (${h=1, 2, \ldots, H}$):

\begin{equation}
    \alpha_{ij}^{(h)} = \text{softmax}((\mathbf{W}^Q_{(h)} \mathbf{e}_i)^\top \mathbf{W}^K_{(h)} \mathbf{e}_j - \lambda^{(h)} d^{2}_{ij})
\end{equation}

\subsection{Computational Pipeline of GA-sklearn}
The proposed computational pipeline for training a GA model and making inference on new points is illustrated in Figure 1. We introduce a redesigned data loading factory in this paper for efficient data preparation and sampling. The factory maintains two distinct sets of data, the \textit{context pool} and the \textit{query pool}. The query pool contains all points to be inferred, which may include training or test data, depending on different scenarios. In contrast, the context pool comprises all observed points that serve as contextual information for the query points. To construct an input sequence, the factory draws one target point from the query pool and queries its spatial neighbors from the context pool. The input sequence is then constructed by combining the target point and its neighbors.

Although GA does not rely on an explicit graph structure to capture spatial context, spatial querying remains essential for constructing input sequences. This is typically accomplished by building a k-d tree using the coordinate features of points in the context pool. The k-d tree is then used to retrieve spatial neighbors of the target points during both training and inference. We train the GA model for multiple epochs in the training stage, and make multiple inferences for model ensembling. Consequently, these processes involve remarkably redundant k-d tree query operations if the neighborhood relationship is calculated on-the-fly. To this end, our implementation pre-computes the spatial neighbors for each query point in advance of these two stages. Specifically, after being provided with a new query pool, the data loading factory retrieves and caches the corresponding neighbors from the context pool.

In the inference stage, we apply model ensembling to further enhance the performance of a single prediction. This is achieved by introducing randomness to each prediction. Specifically, we always expand the searching radius by an expansion factor, making the number of retrieved contextual points slightly exceed the maximum input length. The data loading factory then randomly removes redundant points to construct the actual input sequence. Consequently, for a single target point, its contextual points vary in each inference, resulting in varied predictions.

\subsection{Model Explainability with GeoShapley}
Explainability methods offer critical insights into the decision-making process and the reliability of machine learning models in geospatial analysis. In this paper, we use GeoShapley for this purpose because of its model-agnostic nature. To accommodate different model types (e.g., trees, neural networks), GeoShapley extends the Kernel SHAP (\cite{NIPS2017_7062}) to estimate Shapley values for spatial and non-spatial features so to provide measures of importance in the model. In essence, GeoShapley treats the spatial features (e.g., 2D coordinates or high-dimensional positional embedding features) as one joint feature. Under the Kernel SHAP framework, the Shapley value of each feature, the joint spatial feature and the interaction between spatial and non-spatial features are decomposed into four components.

\begin{equation}
    \hat{\mathbf{y}} = \phi_0 + \boldsymbol{\phi_{GEO}} + \sum^p_{j=1}\boldsymbol{\phi_j} + \sum^p_{j=1}\boldsymbol{\phi_{(GEO,j)}}
\end{equation}
where ${n}$ is the number of samples, ${p}$ is the number of features. ${\phi_0}$ is a constant base value that is the averaged prediction of all background points. ${\boldsymbol{\phi_{GEO}}}$ is a vector of size ${n}$ that represents the intrinsic location effect captured by the model. ${\boldsymbol{\phi_j}}$ is a vector of size ${n}$ that represents the effect of the ${j}$-th non-spatial feature. And finally, ${\boldsymbol{\phi_{(GEO,j)}}}$ is a vector of size ${n}$ representing each non-spatial feature ${j}$ giving the spatially varying interaction effect.

Although the Kernel SHAP approach used in GeoShapley is interoperable with GA, it can be computationally expensive, as it requires a large number of repeated forward passes through the model with permuted inputs. Our redesigned pipeline to enhance model efficiency also enable practical post-hoc model explanations. Specifically, we develop a wrapper method, \verb|get_shap_predictor|, apart from the standard \verb|predict| method, to serve as the predictive interface passed to a standard post-hoc explainer such as the GeoShapley explainer. In our implementation, the data loading factory assigns each point in the dataset a unique ID, so that even when location features are perturbed, we can still reliably retrieve its true neighbors and construct the correct input sequence.

\begin{figure}
    \centering
    \includegraphics[width=1\linewidth]{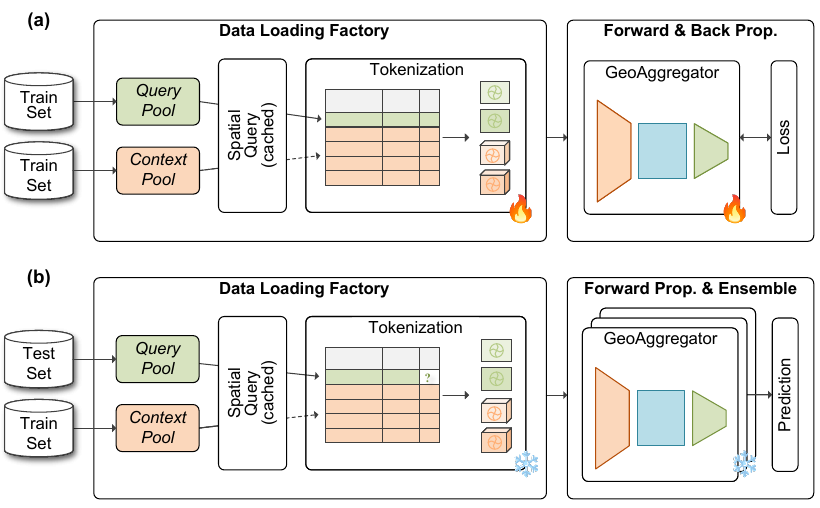}
    \caption{The supervised training (a) and inference (b) pipelines of the GeoAggregator model.}
    \label{fig:enter-label}
\end{figure}

\section{EXPERIMENTS AND RESULTS}
In this section, we apply the redesigned computational pipeline for the GA model on synthetic GTDs to demonstrate its improved accuracy, efficiency and explainability.

\subsection{Experimental Setup}
\textbf{Data.} For the consistency in comparison, we use eight synthetic datasets proposed in \cite{deng2025geoaggregator} in the accuracy assessments. We then select two out of eight for the purpose of evaluating the efficiency and explainability. The first is SL-r, which is characterized by incorporating the SA effect between spatial target variables following a spatial lag model design. The second is GWR-r, which contains a target variable that is linearly combined from two random variables using two corresponding spatially varying coefficients. We then evaluate the explainability of GA on the second dataset. The detailed specifications of the datasets can be found in \cite{deng2025geoaggregator}.

\textbf{Compared Models.} In the evaluation of the explanatory results of GA and given the limited length of the paper, we select one baseline model, XGBoost, which is a strong tabular learner with notable explainability \cite{li2022extracting, shwartz2022tabular}.

\textbf{Experimental Details.} To simulate a real-world usage case of the pipeline, we conduct all the experiments on a laptop with an Intel(R) Core(TM) i7-11390H CPU and 16 GB RAM. In the explanatory experiments, we calculate the GeoShapley values using 30 randomly selected samples as background points.

\subsection{Model Performance}
We first demonstrate the impact of the proposed ensemble mechanism on the accuracy of the regression in Table 1. Without further training, solely applying model ensembling during inference always yields equal or better results on all 8 datasets. Including more ensemble members generally improves the results, but with diminished marginal gain. Hence, users can select the appropriate number of ensemble members in practice to balance the desired performance and the time budget. Moreover, the epistemic uncertainty of each prediction can be approximated by the variance of outputs across ensemble members \cite{lakshminarayanan2017simple}.

\begin{table}
    \centering
    \begin{tabular}{ccc}
    \toprule
    Ensemble Member & ${R^2}$ ($\uparrow$) & MAE ($\downarrow$) \\
    \midrule
        1 & 0.841 & 1.149 \\
        4 & 0.844 & 1.141 \\
        8 & 0.844 & 1.135 \\
    \bottomrule
    \end{tabular}
    \caption{Average performance of GA across 8 synthetic datasets using different numbers of ensemble members.}
    \label{tab:my_label}
\end{table}

\subsection{Model Efficiency}
In this section, we demonstrate the improved efficiency of the redesigned pipeline, particularly during the inference process. Without loss of generality, we use the SL-r dataset containing 2,500 points and divide it into training and test sets with a ratio of 7:3. The model is then trained for 30 epochs, before making inferences on the test set. Note that we allow 8 inference results to be assembled to simulate a real usage scenario. We gradually increase the length of the input sequence and record the corresponding time for the inference processes.

The elapsed wall time is compared with the original implementation in Figure 2. By introducing a pre-computing mechanism, the pipeline avoids the redundant k-d tree query operations, especially during repeated forward calculations of the model. The pipeline exhibits an average efficiency improvement of 36\%. On the other hand, GA variants use learnable induced points to lower the quadratic scaling speed with increasing length of input sequence to linear. As a result, we observe near-linearly growing time curves with increasing length of the input sequence, similar to \cite{deng2025geoaggregator}.

\begin{figure}
    \centering
    \includegraphics[width=0.9\linewidth]{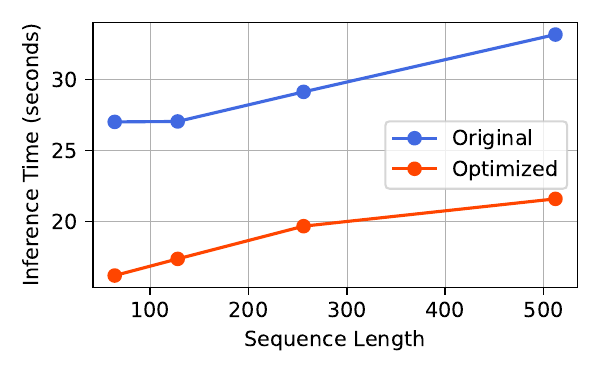}
    \caption{Elapsed inference time of GA with increasing input sequence length, on the SL-r dataset. The dataset contains 2,500 points in total, and the inferences are made on 30\% (750) of the points for 8 times.}
    \label{fig:enter-label}
\end{figure}

\subsection{Model Explanation}
The explainability of GA is illustrated using the GeoShapley framework on the GWR-r dataset and we compare against XGBoost. With this setup, the batch inference strategy manages to generate explanations for GA in approximately 1,800 seconds, while explaining XGBoost takes around 100 seconds. Further work can develop a GPU-friendly version of GeoShapley to better accommodate deep learning-based models. We then compare the true regression coefficients (${\beta}$s) and the retrieved results of XGBoost and GA through GeoShapley in Figure 3. The spatially varying regression coefficients are captured by both models compared to the true distributions. However, the results of XGBoost contain more noisy points and exhibit signatures of hard decision boundaries compared to those of GA. This indicates that GA is an efficient learner to capture the SH effect in GTDs through a localized attention mechanism. 

\begin{figure}
    \centering
    \includegraphics[width=1\linewidth]{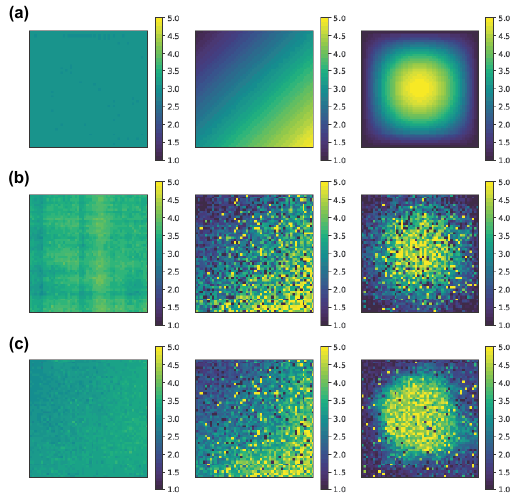}
    \caption{The true spatially varying regression coefficients (a) and the retrieved regression coefficients from GeoShapley for XGBoost (b) and GA (c).}
    \label{fig:enter-label}
\end{figure}

\section{CONCLUSION}
Modeling and explaining GTD are essential tasks in geospatial analysis. An recent proposed transformer-based model, GeoAggregaotor, has demonstrated its promising performance regarding various synthetic and empirical datasets. In this short paper, we further improve GeoAggregator in terms of computational efficiency and model explainability. Accordingly, we propose an optimized computational pipeline that streamlines and speeds up the training and inferring processes of GA. The pipeline features a data loading factory that pre-computes and caches the neighboring relationships of target points and contextual points. We also integrate strategies such as model ensembling to further boost the performance and provide epistemic uncertainty estimation of the regressive results. To explain the result of the model, we provide a wrapper method to enable GeoShapley-based explanation of the model. Experiments show that our designs improve the performance of GA in terms of predictive accuracy and speed. The explanation results further demonstrate that GA can more effectively capture spatial effects compared to XGBoost, indicating it is a stronger spatial learner. 

\bibliographystyle{ACM-Reference-Format}
\bibliography{sample-base}

\end{document}